\documentclass{article}

\usepackage{arxiv}

\usepackage[utf8]{inputenc} % allow utf-8 input
\usepackage[T1]{fontenc}    % use 8-bit T1 fonts
\usepackage{hyperref}       % hyperlinks
\usepackage{url}            % simple URL typesetting
\usepackage{booktabs}       % professional-quality tables
\usepackage{amsmath,amsfonts,amssymb}
\usepackage{nicefrac}       % compact symbols for 1/2, etc.
\usepackage{microtype}      % microtypography
\usepackage{lipsum}
\usepackage{graphicx}
\graphicspath{ {./images/} }
\usepackage{bbm}
\usepackage{array}
\usepackage{hyperref}
\usepackage{algorithmic}
\usepackage[ruled, vlined]{algorithm2e}
\usepackage{color}
\usepackage{stfloats}

\usepackage{colortbl}
\newcolumntype{P}[1]{>{\centering\arraybackslash}p{#1}}
\newcolumntype{R}[1]{>{\flushrifht\arraybackslash}p{#1}}

\definecolor{c_diag}{rgb}{0.85,0.89,0.953}
\definecolor{t_diag}{rgb}{0,0,0}
\definecolor{c_ndiag}{rgb}{.95,0.95,0.95}

\definecolor{c_hlight}{rgb}{.886,0.9411,0.851}
\definecolor{c_hdark}{rgb}{.663,0.82,0.557}
\definecolor{c_rest}{rgb}{.984,0.898,0.839}

\newtheorem{definition}{Definition}

\newtheorem{example}{Example}

\title{The Certainty Ratio $\mathcal{C_\rho}$: a novel metric for assessing the reliability of classifier predictions}

\author{
 Jes\'us S. Aguilar--Ruiz \\
  School of Engineering\\ Pablo de Olavide University\\ 
  ES-41013 Seville, Spain \\
  \texttt{aguilar@upo.es} \\
 }

\begin{document}
\maketitle
\begin{abstract}
Evaluating the performance of classifiers is critical in machine learning, particularly in high--stakes applications where the reliability of predictions can significantly impact decision--making. Traditional performance measures, such as accuracy and F--score, often fail to account for the uncertainty inherent in classifier predictions, leading to potentially misleading assessments. This paper introduces the Certainty Ratio ($C_\rho$), a novel metric designed to quantify the contribution of confident (certain) versus uncertain predictions to any classification performance measure. By integrating the Probabilistic Confusion Matrix ($CM^\star$) and decomposing predictions into certainty and uncertainty components, $C_\rho$ provides a more comprehensive evaluation of classifier reliability. Experimental results across 21 datasets and multiple classifiers, including Decision Trees, Naïve--Bayes, 3--Nearest Neighbors, and Random Forests, demonstrate that $C_\rho$ reveals critical insights that conventional metrics often overlook. These findings emphasize the importance of incorporating probabilistic information into classifier evaluation, offering a robust tool for researchers and practitioners seeking to improve model trustworthiness in complex environments.

\end{abstract}

% keywords can be removed
%\keywords{First keyword \and Second keyword \and More}

\section{Introduction} \label{sec:introduction}

Classification is a fundamental task in machine learning, where the goal is to assign a class label to each instance in a dataset based on input features. A wide range of classification algorithms have been developed and extensively used across various fields, including healthcare, finance, and scientific research. Evaluating the performance of these classifiers is crucial for understanding their effectiveness and reliability in practical applications.

Traditionally, classification performance is assessed using confusion matrices and derived measures such as accuracy, precision, recall, F--score, and the Matthews correlation coefficient, among others. These measures, however, often rely on hard classification decisions, where each instance is definitively assigned to a single class. This approach overlooks the probabilistic nature of many classification models that provide class membership probabilities rather than definitive labels. As a result, conventional confusion matrices may overestimate the performance of classifiers by ignoring the inherent uncertainty in predictions.

This paper introduces the concept of the Probabilistic Confusion Matrix ($CM^\star$), which directly incorporates the probability outputs from classifiers into performance evaluation. By decomposing the probabilistic predictions into components representing certainty and uncertainty, the $CM^\star$ provides a more realistic assessment of classifier performance. The proposed approach highlights the impact of uncertain predictions on commonly used performance metrics and offers new insights into classifier reliability. The certainty ratio ($C_\rho$) quantifies the extent to which a classifier’s performance is driven by confident, reliable predictions versus uncertain ones, and provides a deeper insight into classifier behavior, revealing the proportion of accurate predictions that the model is genuinely confident about. This measure is particularly valuable in high--stakes fields where the cost of uncertain or incorrect predictions can be substantial. By incorporating the certainty ratio into performance evaluation, we can more accurately assess the true reliability of classifiers, guiding the development and selection of models that not only perform well but also inspire greater trust in their predictions.

In this work, we present a comprehensive analysis of the probabilistic confusion matrix across multiple classification methods and datasets. We compare traditional performance measures with those derived from the $CM^\star$ to illustrate the effects of uncertainty on evaluation metrics. Our results emphasize the importance of considering probabilistic information in performance assessment and demonstrate that commonly used accuracy metrics can be misleading when uncertainty is not adequately accounted for.

The remainder of this paper is organized as follows: Sections \ref{sec:definitions}, \ref{sec:confusion_matrix}, and \ref{sec:probabilistic} provide formal definitions of the confusion matrix, prediction matrix, and the probabilistic confusion matrix. Sections \ref{sec:uncertainty} introduces the decomposition of probabilistic predictions into certainty and uncertainty components. The concept of probabilistic measure, and the definition and properties of the certainty ratio is introduced in Section \ref{sec:measures}. Section \ref{sec:experiments} describes the experimental setup and datasets used in this study, so as the analysis of results provided by four classifiers. Section \ref{sec:conclusions} discusses the implications of using probabilistic measures, and concludes with potential directions for future research.

\section{Definitions}\label{sec:definitions}

Let $D=(E,F,\upsilon,\omega)$ be a data set, where $E$ is the set of example (or instance) identifiers, $F$ is the set of feature (variable) identifiers, $\upsilon:E\times F \rightarrow \mathbb{R}$ is the function that assigns a real value to a pair $(e,f)$, where $e\in E$ and $f\in F$. Within the field of supervised learning, when $\omega:E \rightarrow \mathbb{L}$ assigns a class label $c$ to an example $e$, with $\mathbb{L}=\{c_1,\dots,c_k\}$, then it is a classification problem; otherwise, if $\mathbb{L}\subseteq\mathbb{R}$, then it is a regression problem. In absence of example identifiers (e.g. patient identifiers) or feature identifiers (e.g. gene names), it is convenient to use $E=\{e_1,\dots,e_n\}$ and $F=\{f_1,\dots,f_m\}$, respectively. Henceforth, $\upsilon(e_i,f_j)$ will represent a real value, and $\omega(e_i)$ will represent a class label (e.g. type of tumor). 

Both the function $\upsilon$ and the function $\omega$ are expressed in tabular form, so that $\upsilon$ is determined by a $n\times m$ matrix of real values, and $\omega$ as a vector of $n$ values from $\mathbb{L}$. In order to simplify the notation, a function is defined to extract the values of sample $e_i$: $input: E \rightarrow \mathbb{R}^m$, such that $input(e_i)=(\upsilon(e_i,f_1),\dots,\upsilon(e_i,f_m))$.

The main goal of any classification problem is to find a general function $\Omega: \mathbb{R}^m \rightarrow \mathbb{L}$, learned from $D$, such that $\Omega(input(e_i)) = \omega(e_i)$, $\forall e_i$, where $i=\{1,\dots,n\}$, or at least it maximizes the frequency of the equality.

However, the function $\omega$ only extracts the class label from training examples, so $\Omega$ will assign a class label to any test example $e_t$ based on the knowledge learned from training data, and for that, it will use the predictive model $\rho$ built by the classifier. 

\section{Confusion Matrix}\label{sec:confusion_matrix}

\begin{definition}[Dataset] A dataset $D$ can be represented as a $n \times (m+1)$ matrix, with $n$ rows (instances), $m$ columns (input variables), and an additional column for the class label (output variable).
\end{definition}

\begin{example}[Dataset]
For  example, let a dataset $D$ be with six instances, $m$ variables, and three labels \{A, B , C\}.
\begin{equation}
D_{6,m+1}=
\begin{bmatrix} 
 & \cdots &  & A\\
 & \cdots &  & A\\
 & \cdots &  & A\\
 & \cdots &  & B\\
 & \cdots &  & B\\
 & \cdots &  & C\\
\end{bmatrix}
\end{equation}
\end{example}

\begin{definition}[Ground Truth Matrix] 
Let $T_{k,n}$ be the \textit{Ground Truth Matrix} that contains the class label of each example from the dataset $D$. $T_{n,k}$ can be defined as a matrix of $n$ rows (instances) and $k$ columns (classes), where each row represents the class label expressed as a one--hot vector.
\begin{equation}
T_{n,k}=\begin{bmatrix} 
t_{1,1} & \dots & t_{1,k} \\ 
\vdots &  & \vdots \\
t_{n,1} & \dots & t_{n,k} 
\end{bmatrix}
\end{equation}
\noindent where
\begin{equation}
{t_{i,j}} = 
\begin{cases}
1,	&{\text{if}}\ \omega(e_i) = c_j \\ 
{0,}	&{\text{otherwise}} 
\end{cases}
\end{equation}
\end{definition}

\begin{example}[Ground Truth Matrix]
If a dataset has three labels \{A, B , C\}, and an example $e=\{v_1,\dots, v_m, A\}$, then $t_i=\{1,0,0\}$, i.e., the class $A$ is active (1) and the others are inactive (0).

The label column \{A, A, A, B, B, C\} could be transformed into $T_{6,3}$ by means of label binarization, such as:
\begin{equation*}
T_{6,3}=
\begin{bmatrix}
1 & 0 & 0 \\
1 & 0 & 0 \\
1 & 0 & 0 \\
0 & 1 & 0 \\
0 & 1 & 0 \\
0 & 0 & 1 \\
\end{bmatrix}
\end{equation*}
\noindent where each instance is represented in rows, and each class in columns, respectively.
\end{example}

\begin{definition}[Prediction Matrix]
Let $P_{n,k}$ be the \textit{Prediction Matrix} composed of each element $p_{i,j}$ such as:
\begin{equation}
P_{n,k}=\begin{bmatrix} 
p_{1,1} & \dots & p_{1,k} \\ 
\vdots &  & \vdots \\
p_{n,1} & \dots & p_{n,k} 
\end{bmatrix}
\end{equation}
\noindent where $p_{i,j}$ is defined as:
\begin{equation}
{p_{i,j}} = 
\begin{cases}
1,&{\text{if}}\ \Omega(e_i) = c_j \\ 
{0,}&{\text{otherwise}} 
\end{cases}
\end{equation}
\end{definition}

\begin{example}[Prediction Matrix]
A possible example of $P$, given the outputs of a model $\rho$, for the dataset $D$ described earlier, would be:
\begin{equation*}
P_{6,3}=
\begin{bmatrix} 
1 & 0 & 0\\
1 & 0 & 0\\
1 & 0 & 0\\
1 & 0 & 0\\
0 & 1 & 0\\
0 & 1 & 0\\
\end{bmatrix}
\end{equation*}
\noindent where each row refers to an example, i.e., the model $\rho$ classified the first 4 examples as class A, the next 2 examples as class B, and 0 examples as class C.
\end{example}

\begin{definition}[Confusion Matrix]
Let $CM_{k,k}$ be the \textit{Confusion Matrix}, which can be defined as:
\begin{equation}
{CM}_{k,k}=T_{n,k}^\intercal P_{n,k}
\end{equation}
\noindent where $\forall$ $\alpha_{i,j}\in {CM}_{k,k}$, $\alpha_{i,j}\in \mathbb{Z}^+$, i.e., all the values in $CM$ are positive integers.
\end{definition}

\begin{example}[Confusion Matrix]
In the example, multiplying the transposed of $T_{6,3}$, $T_{6,3}^\intercal$, and $P_{6,3}$, will result in ${CM}_{3,3}$.
\begin{equation*}
\small
{CM}_{3,3}=
\begin{bmatrix}
1 & 1 & 1 & 0 & 0 & 0 \\
0 & 0 & 0 & 1 & 1 & 0 \\
0 & 0 & 0 & 0 & 0 & 1 \\
\end{bmatrix}
\begin{bmatrix} 
1 & 0 & 0\\
1 & 0 & 0\\
1 & 0 & 0\\
1 & 0 & 0\\
0 & 1 & 0\\
0 & 1 & 0\\
\end{bmatrix}
=
\begin{bmatrix} 
3 & 0 & 0 \\
1 & 1 & 0 \\
0 & 1 & 0 \\
\end{bmatrix}
\end{equation*}
\end{example}

From ${CM}_{k,k}$ any classification performance measure could be calculated. For instance, the accuracy, $Acc = \frac{4}{6} = 0.67$. 

$T$ is a binary matrix, i.e., all the values $t_{i,j}$ are in \{0,1\}. $P$ is a positive integer matrix, i.e., all the values $p_{i,j}$ are in $\mathbb{Z}^+$. The key aspect is that the confusion matrix $CM$ can be expressed as the product of the transpose of the ground truth matrix $T$ and the prediction matrix $P$. At this point, an interesting question arises: \textbf{what if the prediction matrix $P$ is a positive real matrix (instead of a positive integer matrix), i.e., all the values  $p_{i,j}$ are in $\mathbb{R}^+$?}

\section{Probabilistic Confusion Matrix}\label{sec:probabilistic}

A predictive model $\rho$ usually provides estimates for a test sample $e_t$ to belong to any of the classes, i.e., $\rho(e_t)=\{q_1,\dots, q_k\}$, where $q_j$ are probabilities, i.e., $q_j \in [0,1]$ and $\sum_j q_j = 1$. Then, the function $\Omega$ is defined as  $\Omega_\rho(e_t)=\{c_j | j=\arg \max_{j} \rho(e_t)\}$, where $\rho$ could be any classification technique (e.g., decision trees, nearest neighbors, neural networks, etc.).

The prediction matrix $P_{n,k}$ is then built by binarizing the outputs of $\rho$. However, we could use the probabilities directly in that matrix. Therefore, we could define a new prediction matrix, $Q_{n,k}$, whose values are directly obtained from the classifier $\rho$ (from which the matrix $P_{n,k}$ is calculated).

\begin{definition}[Probabilistic Prediction Matrix]
The matrix $Q_{n,k}$ is obtained directly from the classifier $\rho$, in such a way that each row represents the output probabilities of $\rho$ for each class, i.e., $q_i = \rho(e_i)$.
\begin{equation}
Q_{n,k}=\begin{bmatrix} 
q_{1,1} & \dots & q_{1,k} \\ 
\vdots &  & \vdots \\
q_{n,1} & \dots & q_{n,k} 
\end{bmatrix}
\end{equation}
\end{definition}

\begin{example}[Probabilistic Prediction Matrix]
A possible example of $Q_{6,3}$, given the outputs of $\rho$ for the dataset $D$, would be:
\begin{equation*}
Q_{6,3}=
\begin{bmatrix} 
0.9 & 0.1 & 0\\
0.8 & 0 & 0.2\\
0.6 & 0.1 & 0.3\\
0.4 & 0.3 & 0.3\\
0.1 & 0.8 & 0.1 \\
0 & 0.9 & 0.1 \\
\end{bmatrix}
\end{equation*}
In principle, since $P_{6,3}$ was originally derived from $Q_{6,3}$, the information contained in $Q$ is richer than that in $P$. However, most classification performance measures are based on the $CM$, which is directly obtained from $P$ (except for those based on probabilities, such as, for instance: the Brier score \cite{Brier1950}, the ROC curve \cite{Peterson1954, Swets1973, Metz1978, Hanley1982, Hand2001, Fawcett2006}, and the MCP curve \cite{Aguilar2022a}). The potential of using $Q$ instead of $P$ will be analyzed next.
\end{example}

\begin{definition}[Probabilistic Confusion Matrix]
Let $CM^\star_{k,k}$ be the \textit{Probabilistic Confusion Matrix}, which can be defined as:
\begin{equation}
{CM^\star}_{k,k}=T_{n,k}^\intercal Q_{n,k}
\end{equation}
\noindent where $\forall$ $\beta_{i,j}\in {CM^\star}_{k,k}$, $\beta_{i,j}\in \mathbb{R}^+$, i.e., all the values in $CM^\star$ are positive reals.

Henceforth, all the probability--based measures will be expressed using the symbol $^\star$ as a superscript.
\end{definition}

\begin{example}[Probabilistic Confusion Matrix]
In the previous example, multiplying the transposed of $T_{6,3}$, $T_{6,3}^\intercal$, and $Q_{6,3}$, will be obtained ${CM^\star}_{3,3}$.
%\begin{equation*}
\begin{align*}
{CM^\star}_{3,3}=  
& \begin{bmatrix}
1 & 1 & 1 & 0 & 0 & 0 \\
0 & 0 & 0 & 1 & 1 & 0 \\
0 & 0 & 0 & 0 & 0 & 1 \\
\end{bmatrix}
\begin{bmatrix} 
0.9 & 0.1 & 0\\
0.8 & 0 & 0.2\\
0.6 & 0.1 & 0.3\\
0.4 & 0.3 & 0.3\\
0.1 & 0.8 & 0.1 \\
0 & 0.9 & 0.1 \\
\end{bmatrix} = 
\begin{bmatrix} 
2.3 & 0.2 & 0.5 \\
0.5 & 1.1 & 0.4 \\
0 & 0.9 & 0.1 \\
\end{bmatrix}
\end{align*}
%\end{equation*}
${CM^\star}_{3,3}$ is slightly different from ${CM}_{3,3}$. All the rows from both matrices must sum up the number of examples from each class \{A=3, B=2, C=1\}. However, the column sum, which indicates the precision for each class, is different. For class A, in $CM$ is 3, and in $CM^\star$ is 2.8.
\end{example}

\section{Uncertainty Matrix}\label{sec:uncertainty}

The probabilistic confusion matrix $CM^\star$ provides information about the classification probabilities, indicating the accurate of predictions made by $\Omega$ for each test example based on the probabilities generated by the classification model $\rho$. However, not all probabilities are computed equally because some correspond to correct classifications and others to incorrect classifications. Therefore, it is possible to decompose $CM^\star$ into two matrices to analyze the degree of certainty and uncertainty of the predictions.

\begin{definition}[Decomposition of $Q$]
The matrix $Q_{n,k}$ can be decomposed into two matrices, such that $Q_{n,k} = Q^+_{n,k} + Q^-_{n,k}$, and, for all $q^+ \in Q^+$ and $q^- \in Q^-$:
\begin{equation}
{q^+_{i,j}}=
\begin{cases}
q_{i,j},&{\text{if}}\ q_{i,j} = max_i q_{i,j} \\ 
{0,}&{\text{otherwise}} 
\end{cases}
\end{equation}
\begin{equation}
{q^-_{i,j}}=
\begin{cases}
q_{i,j},&{\text{if}}\ q_{i,j} \neq max_i q_{i,j} \\ 
{0,}&{\text{otherwise}} 
\end{cases}
\end{equation}
\end{definition}

The goal of this decomposition is to separate $Q_{n,k}$ into two matrices, $Q^+_{n,k}$ and $Q^-_{n,k}$, where $Q^+_{n,k}$ contains the probability values that lead to correct classifications, and  $Q^-_{n,k}$ contains the probability values that lead to incorrect classifications. 

\begin{example}[Decomposition of $Q$]
\begin{equation*}
\small
\begin{bmatrix} 
0.9 & 0.1 & 0\\
0.8 & 0 & 0.2\\
0.6 & 0.1 & 0.3\\
0.4 & 0.3 & 0.3\\
0.1 & 0.8 & 0.1 \\
0 & 0.9 & 0.1 \\
\end{bmatrix}
=
\begin{bmatrix} 
0.9 & 0 & 0\\
0.8 & 0 & 0\\
0.6 & 0 & 0\\
0.4 & 0 & 0\\
0 & 0.8 & 0 \\
0 & 0.9 & 0 \\
\end{bmatrix}
+
\begin{bmatrix} 
0 & 0.1 & 0\\
0 & 0 & 0.2\\
0 & 0.1 & 0.3\\
0 & 0.3 & 0.3\\
0.1 & 0 & 0.1 \\
0 & 0 & 0.1 \\
\end{bmatrix}
\end{equation*}
The matrix $Q^+_{n,k}$ contains the highest probability values for each row, representing the probabilities that were decisive in providing the matrix $P$. For instance, these values are 0.9 (first row), 0.8 (second row), 0.6 (third row), 0.4 (fourth row), 0.8 (fifth row), and 0.9 (sixth row). In contrast, the matrix $Q^-_{n,k}$ contains the remaining values that did not correspond to the highest probability in each row, reflecting the uncertainty in the predictions. 
\end{example}

It is important to emphasize that in the confusion matrix $CM$, only the values from $Q^+_{n,k}$ are considered (transforming values different from 0 into 1), while the values from $Q^-_{n,k}$ are not included. This means that the confusion matrix focuses solely on the probabilities that lead to the most confident predictions, quantitatively disregarding  those associated with lower certainty and incorrect classifications. This selective consideration helps to highlight the model's performance based on its most confident decisions, while the excluded values provide insights into the areas of uncertainty and error.
 
\begin{definition}[Decomposition of $CM^\star$]
Let $CM^\star$ be the probabilistic confusion matrix. $CM^\star$ can be decomposed into two matrices $CM^\star= U + V$, such as $u\in U$ and $v\in V$ are calculated following the next equations.
\begin{equation}
CM^\star_{k,k} = T_{n,k}^\intercal \left(Q^+_{n,k} + Q^-_{n,k}\right) = V_{k,k} + U_{k,k}
\end{equation}
\end{definition}

The matrix $V_{k,k}$ represents the certainty of $CM^\star$, as it is calculated from the values of probabilities assigned to the prediction, and the matrix $U_{k,k}$ represents the uncertainty, as these are values do not correspond to correct classifications.

\begin{example}[Decomposition of $CM^\star$]\label{ex:decomposition}
\small
\begin{equation*}
\begin{bmatrix} 
2.3 & 0.2 & 0.5 \\
0.5 & 1.1 & 0.4 \\
0 & 0.9 & 0.1 \\
\end{bmatrix}=
\begin{bmatrix} 
2.3 & 0 & 0 \\
0.4 & 0.8 & 0 \\
0 & 0.9 & 0 \\
\end{bmatrix}+
\begin{bmatrix} 
0 & 0.2 & 0.5 \\
0.1 & 0.3 & 0.4 \\
0 & 0 & 0.1 \\
\end{bmatrix}
\end{equation*}
\end{example}

\begin{definition}[Certainty Matrix ($V$)]
The matrix $V$ collects the contribution to the $CM^\star$ of potentially correct predictions made by the classifier $\rho$. It represents the certainty of the classifier, as it assigns a class to a test example with higher probability than to other classes. From the matrix $V$ any probabilistic measure can be calculated. 
\end{definition}

\begin{definition}[Uncertainty Matrix (U)]
The matrix $U$ collects the contribution to the probabilistic confusion matrix $CM^\star$ from potentially incorrect predictions made by the classifier $\rho$. It represents the uncertainty of the classifier, as the probabilities present in this matrix correspond to classes different from the predicted class of $\rho$. From the matrix $U$, various probabilistic measure can be calculated. 
\end{definition}
 
 \section{Probabilistic measures}\label{sec:measures}
 
 There exist a number of measures to assess the classification performance (accuracy, precision, recall, $F_\beta$--score, Matthews correlation coefficient, etc.). All of these measures are based on the confusion matrix $CM_{k,k}$.
 
 \begin{definition}[Classification performance measure]
 A classification performance measure is a function $\psi$ defined as follows:
 \begin{equation}
 \psi : \mathbb{R}^{k \times k}  \longrightarrow \mathbb{R} \quad \mbox{such that } \psi(CM_{k,k}) = r \in \mathbb{R}
 \end{equation}
  \end{definition}
  
\begin{example}[Accuracy]
Accuracy is a type of function $\psi$ defined as: 
\begin{equation*}
 Acc = \frac{\sum_{i=1}^k \alpha_{i,i}}{\sum_{i,j=1,1}^{k,k} \alpha_{i,j}}
 \end{equation*}
 \noindent where $k$ is the number of classes (number of rows or columns of the matrix), and $\forall i,j \in \{1,\dots,k\}$, $\alpha_{i,j} \in CM_{k,k}$.
\end{example}
 
 \begin{definition}[Probabilistic classification performance measure]
 A probabilistic classification performance measure is a function $\phi$ defined as follows:
 \begin{equation}
  \phi : \mathbb{R}^{k \times k}  \longrightarrow \mathbb{R} \quad \mbox{such that }  \phi(CM^\star_{k,k}) = s \in \mathbb{R}
 \end{equation}
 Any classification performance measure $\psi$ can be expressed as a probabilistic classification performance measure $\phi$ by simply replacing the confusion matrix $CM_{k,k}$ with the probabilistic confusion matrix $CM^\star_{k,k}$.
\end{definition}
 
\begin{example}[Accuracy]\label{ex:accuracy}
From the $CM^\star$ matrix, the accuracy $Acc^\star= \frac{3.5}{6}=0.583$. 
\end{example}
 
 \begin{definition}[Decomposition of probabilistic classification performance measure]
 Any probabilistic classification performance measure can be decomposed into two measures: one obtained from the certainty matrix $V$ and one obtained from the uncertainty matrix $U$. Since the probabilistic confusion matrix can be expressed as $CM^\star_{k,k}=V_{k,k}+U_{k,k}$, the function $\phi$ can be decomposed into two functions $\phi_v$ and $\phi_u$, corresponding to the certainty and uncertainty contributions, respectively, as follows: 
\begin{equation}
\phi(CM^\star_{k,k}) = \tau\left(\phi_v(V_{k,k}),\phi_u(U_{k,k})\right)
\end{equation}
\noindent where $\tau$ is a function.
 \end{definition}
 
 \begin{example}[Probabilistic Accuracy]
 \begin{equation}
Acc^\star = \lambda_v Acc^\star_v + \lambda_u Acc^\star_u
\end{equation}
\noindent where
\begin{equation}
\lambda_v =  \frac{\sum_{i,j=1,1}^{k,k} v_{i,j}}{\sum_{i,j=1,1}^{k,k} (u_{i,j}+v_{i,j})}
\end{equation}
\begin{equation}
\lambda_u = \frac{\sum_{i,j=1,1}^{k,k} u_{i,j}}{\sum_{i,j=1,1}^{k,k} (u_{i,j}+v_{i,j})}
\end{equation}
\begin{equation}
Acc^\star_v= \frac{\sum_{i=1}^{k} v_{i,i}}{\sum_{i,j=1,1}^{k,k} v_{i,j}}
\end{equation}
\begin{equation}
Acc^\star_u= \frac{\sum_{i=1}^{k} u_{i,i}}{\sum_{i,j=1,1}^{k,k} u_{i,j}}
\end{equation}
\noindent where $v_{i,j} \in V_{k,k}$, and $u_{i,j} \in U_{k,k}$.
 
In Example \ref{ex:accuracy}, $Acc^\star=0.583$, which can be decomposed as follows:
\begin{equation*}
Acc^\star = \frac{4.4}{6}\cdot\frac{3.1}{4.4}+\frac{1.6}{6}\cdot\frac{0.4}{1.6}=0.73\cdot 0.705 + 0.27\cdot 0.25 = 0.583
\end{equation*}
\end{example}

The coefficients $\lambda_v$ and $\lambda_u$ are extremely important. $\lambda_v$ measures the ratio of probabilities used in the certainty matrix $V$, and $\lambda_u$ measures the ratio of probabilities used in the uncertainty matrix $U$. In the example, $\lambda_v=0.73$, which means that about 73\% of predictions were made with confidence (this does not mean that they were accurate). In contrast, the system assigned about 27\% of probabilities to predictions without confidence (and some might even be accurate by chance). 
 
 \begin{definition}[Probabilistic divergence]
Let $CM$ be the confusion matrix and $CM^\star$ be the probabilistic confusion matrix obtained for a dataset $D$ after applying a classifier $\rho$. The \textit{probabilistic divergence} of $CM^\star$ with respect to $CM$ is defined as follows:
\begin{equation}
d(CM,CM^\star)=\frac{1}{n}\sqrt{\sum_{i=1}^k\sum_{j=1}^k \left(\alpha_{i,j}-\beta_{i,j}\right)^2 }
\end{equation}
\noindent where $n$ is the number of test instances, and $\forall i,j$ $\alpha_{i,j}\in {CM}_{k,k}$ and $\beta_{i,j}\in {CM^\star}_{k,k}$.
\end{definition}

\begin{example}[Probabilistic divergence]
The probabilistic divergence between $CM$ and $CM^\star$, given classifier $\rho$ for the dataset $D$ is:
%\begin{equation}
%\mathrm{confidence}(M)=\frac{1}{6}\left( (0.7)^2+(-0.2)^2+(-0.5)^2+(0.5)^2+(-0.1)^2+(-0.4)^2+(0)^2+(0.1)^2+(-0.1)^2\right)
\fontsize{8}{10}\selectfont
\begin{align*}
d(CM,CM^\star)= \frac{1}{6} \sqrt{ ( (0.7)^2+(-0.2)^2+\dots+(-0.1)^2 )} =\frac{\sqrt{1.22}}{6}= 0.184
\end{align*}
\normalsize

%\end{equation}
\end{example}
Since $0\leq d(CM,CM^\star) \leq 1$, this normalized divergence quantifies how far the confusion matrix $CM$ is from the probabilistic confusion matrix $CM^\star$. In other words, it measures how much, on average, the discrete predictions differ from the probabilistic predictions. In this case, the divergence is about 18.4\%, indicating the level of discrepancy between the two matrices. Ideally, if $CM = CM^\star$ then $d=0$, which implies zero uncertainty, meaning the predictions perfectly align with the probabilistic assessments.

The divergence between $CM$ and $CM^\star$ indicates the amount of uncertainty present in the classification performance provided by the classifier $\rho$. This measure allows us to understand the level of uncertainty in the predictions, regardless of the classification performance measure used or its effectiveness. Therefore, by comparing the divergence values, we can identify which classifier, among multiple options, produces the least amount of uncertainty for a given dataset $D$. This insight helps in selecting classifiers that not only perform well but also offer more reliable and certain predictions.

\begin{definition}[Certainty Ratio]
The certainty ratio $\mathcal{C_\rho}$ of a probabilistic classification performance measure for a classifier $\rho$, that provides a probabilistic confusion matrix $CM^\star_{k,k}$ decomposed into the certainty matrix $V$ and the uncertainty matrix $U$, is defined as follows:
 \begin{equation}
\mathcal{C_\rho} = \frac{\phi_v(V_{k,k})}{\phi_v(V_{k,k}) + \phi_u(U_{k,k})} 
\end{equation}
 \end{definition}
 \begin{example}[Certainty Ratio (Accuracy)]
 Considering that $\phi_v(V_{k,k})=Acc^\star_v$ and $\phi_u(U_{k,k}) = Acc^\star_u$:
 \begin{equation}
 Acc^\star_v = \frac{2.3+0.8}{4.4} =0.705 \quad Acc^\star_u = \frac{0.3+0.1}{1.6} = 0.25
 \end{equation}
 the certainty ratio $\mathcal{C_\rho}$ is:
%\fontsize{8}{10}\selectfont
 \begin{equation*}
\mathcal{C_\rho} = \frac{0.705}{0.705+0.25} = 0.738
 \end{equation*}
 \normalsize

This means that when measuring accuracy  for classifier $\rho$ about 74\% of predictions are attributed to certainty, while 26\% are due to uncertainty.
 \end{example}
 
The certainty ratio $\mathcal{C_\rho}$ is bounded: $0 \leq \mathcal{C_\rho} \leq 1$. $\mathcal{C_\rho} = 1$ implies that all performance is derived from certain predictions, indicating complete confidence in the classifier’s decisions.  $\mathcal{C_\rho} = 0$ suggests that all performance is due to uncertain or incorrect predictions, indicating a lack of reliable decisions. $\mathcal{C_\rho}$ quantifies the proportion of performance attributable to certain predictions, directly linking model evaluation to the confidence of its outputs. This distinguishes it from traditional measures that treat all correct predictions equally, regardless of underlying certainty. In high--stakes applications, a high $\mathcal{C_\rho}$ signals that the classifier’s decisions are robust and trustworthy, while a low $\mathcal{C_\rho}$ serves as a warning that the classifier’s performance is heavily influenced by uncertain predictions. The $\mathcal{C_\rho}$ effectively penalizes classifiers that perform well by chance or due to uncertain decisions, distinguishing between truly reliable models and those whose apparent performance may be inflated by noise or ambiguity. $\mathcal{C_\rho}$ can be particularly valuable in evaluating models trained on noisy or imbalanced datasets, where high traditional accuracy may not reflect genuine predictive power. 

In cases where the classifier consistently assigns high probabilities to the correct classes (ideal conditions), $\phi_v(V_{k,k})  \approx \phi(CM_{k,k}^\star$), and thus  $\mathcal{C_\rho} \approx 1$. This scenario is often desired in medical diagnostics, where confident predictions are crucial for effective decision--making. Conversely, when the classifier frequently assigns low probabilities to the correct class or when the highest probability does not correspond to the correct class, $\phi_u(U_{k,k})$ becomes significant, leading to a lower $\mathcal{C_\rho}$. This situation reflects real--world conditions like overlapping class distributions, ambiguous features, or insufficient training data, where classifier performance cannot be fully trusted. The certainty ratio $\mathcal{C_\rho}$ is sensitive to both the probability distribution across classes and the absolute probability values assigned to the predicted classes. For example, a prediction with a correct class probability of 0.9 will contribute more to $\phi_v(V_{k,k})$, and thus increasing $\mathcal{C_\rho}$, compared to a scenario where the correct class probability is 0.3 and other classes have similar probabilities. This latter situation is more likely when the dataset has a high number of classes.

Calibration techniques \cite{Zadrozny2001,Naeini20152901}, assess how well predicted probabilities match observed frequencies. A well--calibrated model can still have a low $\mathcal{C_\rho}$ if it produces many uncertain predictions, highlighting that calibration alone is not sufficient to guarantee reliable performance. $\mathcal{C_\rho}$ adds value by explicitly measuring the proportion of performance attributable to certainty, complementing calibration metrics in evaluating overall model trustworthiness. In real-time systems, $\mathcal{C_\rho}$ could be used to dynamically adjust classifier thresholds or trigger additional data collection when uncertainty is high, enhancing system robustness and decision quality.

\section{Experimental analysis}\label{sec:experiments}

\begin{table}[t]
\caption{Description of datasets: dataset name, number of samples (\#s), of variables (\#v) and of classes (\#c), respectively.}
\label{tab:datasets}
\begin{center}
\fontsize{8}{10}\selectfont
\begin{tabular}{|l|r|r|r|}
\hline
Dataset							& \#s	& \#v	& \#c \\ \hline \hline
australian credit					& 690		& 14			& 2 \\ \hline
cardiotocography morphologic pattern	& 2,126	& 23	& 10 \\ \hline
%chronic kidney disease full	& 399	& 24	& 2 \\ \hline
customer churn	& 3,150	& 13	& 2 \\ \hline
data banknote authentication	& 1,372	& 4	& 2 \\ \hline
%ecoli	& 336	& 7	& 8 \\ \hline
fertility diagnosis	& 100	& 9	& 2 \\ \hline
hepatitis c	& 1,385	& 28	& 4 \\ \hline
landsat satellite	& 6,435	& 36	& 6 \\ \hline
%leaf	& 340	& 14	& 30 \\ \hline
magic gamma telescope	& 19,020	& 10	& 2 \\ \hline
movement libras	& 360	& 90	& 15 \\ \hline
musk v1	& 476	& 166	& 2 \\ \hline
musk v2	& 6,598	& 166	& 2 \\ \hline
parkinson	& 195	& 22	& 2 \\ \hline
phishing websites	& 11,055	& 30	 & 2 \\ \hline
planning relax	& 182	& 12	& 2 \\ \hline
rice cammeo osmancik	& 3,810	& 7	& 2 \\ \hline
room occupancy	& 10,129	& 16	& 4 \\ \hline
sonar	& 208	& 60	& 2 \\ \hline
speaker accent recognition	& 329	& 12	& 6 \\ \hline
vertebral column	& 310	& 6	& 3 \\ \hline
vowel	& 990	& 13	& 11 \\ \hline
wine quality red	& 1,599	& 11	& 6 \\ \hline
%wine quality white	& 4,898	& 11	& 7 \\ \hline
%yeast	& 1,484	& 8	& 10 \\ \hline  
%Mean		&	--	& --			& --	\\ \hline  
\end{tabular}
\end{center}
\end{table}

\begin{table}[t]
\caption{Results for 3--Nearest Neighbors. }
\label{tab:3NN}
\begin{center}
\fontsize{6.3}{8.4}\selectfont
\begin{tabular}{|l|r|r|r|r|r|r|r|}
\hline
Dataset & $Acc$ & $Acc^\star$ & $Acc_v^\star$ & $Acc_u^\star$ & div & $C_\rho$ & IMCP\\ \hline \hline
australian credit & 0.672 & 0.636 & 0.689 & 0.401 & 6.0 & 63.2 & 0.603 \\ \hline
cardiotocography   & 0.719 & 0.680  & 0.768 & 0.297 & 3.8 & 72.2 & 0.601 \\ \hline
customer churn & 0.866 & 0.843 & 0.886 & 0.360  & 3.6 & 71.3 & 0.687 \\ \hline
data banknote  & 1.000     & 1.000     & 1.000     & 0.000     & 0.0 & 100.0   & 1.000 \\ \hline
fertility diagnosis & 0.880  & 0.813 & 0.880  & 0.117 & 10.4 & 90.7 & 0.547 \\ \hline
hepatitis c & 0.256 & 0.244 & 0.259 & 0.225 & 11.6 & 53.5 & 0.236 \\ \hline
landsat satellite & 0.911 & 0.894 & 0.930  & 0.327 & 1.3 & 74.0 & 0.868 \\ \hline
magic gamma  & 0.801 & 0.770  & 0.822 & 0.363 & 4.6 & 69.4 & 0.716 \\ \hline
movement libras & 0.808 & 0.767 & 0.848 & 0.261 & 5.5  & 77.4 & 0.754 \\ \hline
musk v1 & 0.861 & 0.828 & 0.890  & 0.362 & 4.6 & 71.4 & 0.817 \\ \hline
musk v2 & 0.965 & 0.943 & 0.973 & 0.217 & 3.3 & 81.8 & 0.902 \\ \hline
parkinson & 0.852 & 0.832 & 0.873 & 0.395 & 5.2 & 70.3 & 0.747 \\ \hline
phishing websites & 0.949 & 0.940  & 0.960  & 0.367 & 1.0 & 72.4 & 0.933 \\ \hline
planning relax & 0.622 & 0.579 & 0.630  & 0.411 & 14.7 & 61.8 & 0.472 \\ \hline
rice cammeo  & 0.888 & 0.856 & 0.902 & 0.290  & 3.5 & 75.7 & 0.839 \\ \hline
room occupancy & 0.997 & 0.996 & 0.997 & 0.362 & 0.1 & 76.0 & 0.987 \\ \hline
sonar & 0.821 & 0.803 & 0.843 & 0.398 & 4.1 & 69.6 & 0.785 \\ \hline
speaker accent  & 0.833 & 0.769 & 0.878 & 0.225 & 7.2 & 80.8 & 0.702 \\ \hline
vertebral column & 0.816 & 0.802 & 0.853 & 0.409 & 5.4 & 68.7 & 0.726 \\ \hline
vowel & 0.974 & 0.952 & 0.982 & 0.246 & 2.0 & 81.2 & 0.944 \\ \hline
winequality red & 0.508 & 0.497 & 0.553 & 0.346 & 4.6 & 61.5 & 0.252 \\ \hline \hline
\textbf{Mean}  & \textbf{0.809} & \textbf{0.783} & \textbf{0.829} & \textbf{0.304} & \textbf{4.9} & \textbf{73.5} & \textbf{0.720}\\ \hline
\end{tabular}
\end{center}
\end{table}

\begin{table}[t]
\caption{Results for Naïve-Bayes. }
\label{tab:NB}
\begin{center}
\fontsize{6.3}{8.4}\selectfont
\begin{tabular}{|l|r|r|r|r|r|r|r|}
\hline
Dataset & $Acc$ & $Acc^\star$ & $Acc_v^\star$ & $Acc_u^\star$ & div & $C_\rho$ & IMCP\\ \hline \hline
australian credit & 0.787 & 0.790 & 0.802 & 0.524 & 2.5 & 61.9 & 0.764 \\ \hline
cardiotocography   & 0.592 & 0.589 & 0.607 & 0.373 & 1.4 & 62.0 & 0.656 \\ \hline
customer churn & 0.732 & 0.731 & 0.740  & 0.493 & 0.4 & 60.1 & 0.799 \\ \hline
data banknote  & 0.843 & 0.810  & 0.871 & 0.361 & 3.7 & 70.9 & 0.745 \\ \hline
fertility diagnosis & 0.850  & 0.794 & 0.861 & 0.236 & 10.4 & 81.9 & 0.506 \\ \hline
hepatitis c & 0.240  & 0.248 & 0.241 & 0.251 & 11.2 & 48.6 & 0.290 \\ \hline
landsat satellite & 0.798 & 0.795 & 0.801 & 0.347 & 0.3 & 70.3 & 0.782 \\ \hline
magic gamma  & 0.727 & 0.721 & 0.737 & 0.450  & 2.9 & 62.1 & 0.631 \\ \hline
movement libras & 0.633 & 0.631 & 0.640  & 0.364 & 2.1 & 66.6 & 0.630 \\ \hline
musk v1 & 0.735 & 0.735 & 0.739 & 0.470  & 1.4 & 65.5 & 0.737 \\ \hline
musk v2 & 0.839 & 0.838 & 0.841 & 0.437 & 0.2 & 65.9 & 0.795 \\ \hline
parkinson & 0.713 & 0.709 & 0.715 & 0.525 & 1.4 & 60.5 & 0.771 \\ \hline
phishing websites & 0.604 & 0.616 & 0.613 & 0.726 & 1.7 & 46.0 & 0.662 \\ \hline
planning relax & 0.660  & 0.571 & 0.659 & 0.336 & 23.4 & 66.8 & 0.452 \\ \hline
rice cammeo  & 0.911 & 0.906 & 0.917 & 0.379 & 0.5 & 71.0 & 0.893 \\ \hline
room occupancy & 0.964 & 0.956 & 0.969 & 0.281 & 0.9 & 77.7 & 0.853 \\ \hline
sonar & 0.669 & 0.674 & 0.678 & 0.556 & 2.7 & 56.6 & 0.673 \\ \hline
speaker accent  & 0.574 & 0.546 & 0.627 & 0.225 & 6.1 & 73.8 & 0.561 \\ \hline
vertebral column & 0.823 & 0.790  & 0.854 & 0.341 & 4.7 & 72.0 & 0.688 \\ \hline
vowel & 0.657 & 0.549 & 0.708 & 0.233 & 5.8 & 75.3 & 0.503 \\ \hline
winequality red & 0.546 & 0.473 & 0.565 & 0.293 & 6.3 & 65.9 & 0.293 \\ \hline \hline
\textbf{Mean}  & \textbf{0.709} & \textbf{0.689} & \textbf{0.723} & \textbf{0.390} & \textbf{4.3} & \textbf{65.8} & \textbf{0.652}\\ \hline
\end{tabular}
\end{center}
\end{table}

\begin{table}[t]
\caption{Results for Decision Trees.}
\label{tab:DT}
\begin{center}
\fontsize{6.3}{8.4}\selectfont
\begin{tabular}{|l|r|r|r|r|r|r|r|}
\hline
Dataset & $Acc$ & $Acc^\star$ & $Acc_v^\star$ & $Acc_u^\star$ & div & $C_\rho$ & IMCP\\ \hline \hline
australian credit & 0.851 & 0.851 & 0.851 & 0.000     & 0.0     & 100.0   & 0.848 \\ \hline
cardiotocography   & 0.855 & 0.855 & 0.856 & 0.227 & 0.2 & 86.2 & 0.798 \\ \hline
customer churn & 0.940  & 0.940  & 0.943 & 0.508 & 0.4 & 66.9 & 0.878 \\ \hline
data banknote  & 0.986 & 0.986 & 0.986 & 0.000     & 0.0     & 100.0   & 0.986 \\ \hline
fertility diagnosis & 0.760  & 0.755 & 0.759 & 0.000     & 0.7 & 100.0   & 0.613 \\ \hline
hepatitis c & 0.258 & 0.258 & 0.258 & 0.000     & 0.0     & 100.0   & 0.257 \\ \hline
landsat satellite & 0.861 & 0.861 & 0.861 & 0.000     & 0.0     & 100.0   & 0.839 \\ \hline
magic gamma  & 0.818 & 0.818 & 0.818 & 0.000     & 0.0     & 100.0   & 0.800 \\ \hline
movement libras & 0.706 & 0.706 & 0.706 & 0.000    & 0.0     & 100.0   & 0.715 \\ \hline
musk v1 & 0.771 & 0.771 & 0.771 & 0.000     & 0.0     & 100.0   & 0.77 \\ \hline
musk v2 & 0.971 & 0.971 & 0.971 & 0.000     & 0.0     & 100.0  & 0.944 \\ \hline
parkinson & 0.872 & 0.872 & 0.872 & 0.000     & 0.0     & 100.0   & 0.822 \\ \hline
phishing websites & 0.963 & 0.963 & 0.968 & 0.479 & 0.3 & 67.1 & 0.961 \\ \hline
planning relax & 0.665 & 0.665 & 0.665 & 0.000     & 0.0     & 100.0   & 0.586 \\ \hline
rice cammeo  & 0.876 & 0.876 & 0.876 & 0.000     & 0.0     & 100.0   & 0.873 \\ \hline
room occupancy & 0.995 & 0.995 & 0.995 & 0.000     & 0.0     & 100.0   & 0.985 \\ \hline
sonar & 0.725 & 0.725 & 0.725 & 0.000     & 0.0     & 100.0   & 0.725 \\ \hline
peaker accent  & 0.690  & 0.690  & 0.690  & 0.000     & 0.0     & 100.0   & 0.599 \\ \hline
vertebral column & 0.781 & 0.781 & 0.781 & 0.000     & 0.0     & 100.0   & 0.701 \\ \hline
vowel & 0.819 & 0.819 & 0.819 & 0.000     & 0.0     & 100.0   & 0.819 \\ \hline
winequality red & 0.623 & 0.623 & 0.623 & 0.000     & 0.0     & 100.0   & 0.368 \\ \hline \hline
\textbf{Mean}  & \textbf{0.799} & \textbf{0.799} & \textbf{0.800} & \textbf{0.058} & \textbf{0.0} & \textbf{96.2} & \textbf{0.757}\\ \hline
\end{tabular}
\end{center}
\end{table}

\begin{table}[t]
\caption{Results for Random Forest.}
\label{tab:RF}
\begin{center}
\fontsize{6.3}{8.4}\selectfont
\begin{tabular}{|l|r|r|r|r|r|r|r|}
\hline
Dataset & $Acc$ & $Acc^\star$ & $Acc_v^\star$ & $Acc_u^\star$ & div & $C_\rho$ & IMCP\\ \hline \hline
australian credit & 0.872 & 0.789 & 0.893 & 0.238 & 8.6 & 79.0 & 0.699 \\ \hline
cardiotocography   & 0.895 & 0.769 & 0.924 & 0.124 & 5.8  & 88.2 & 0.625 \\ \hline
customer churn & 0.956 & 0.925 & 0.968 & 0.241 & 3.2 & 80.3 & 0.800 \\ \hline
data banknote  & 0.993 & 0.978 & 0.995 & 0.123 & 1.7 & 90.0 & 0.944 \\ \hline
fertility diagnosis & 0.860  & 0.802 & 0.876 & 0.236 & 11.0 & 81.1 & 0.546 \\ \hline
hepatitis c & 0.258 & 0.252 & 0.260  & 0.248 & 7.5 & 50.9 & 0.292 \\ \hline
landsat satellite & 0.916 & 0.837 & 0.942 & 0.170  & 4.4 & 84.7 & 0.758 \\ \hline
magic gamma  & 0.882 & 0.805 & 0.902 & 0.235 & 8.9 & 79.3 & 0.711 \\ \hline
movement libras & 0.836 & 0.570  & 0.894 & 0.077 & 10.3 & 92.1 & 0.515 \\ \hline
musk v1 & 0.899 & 0.751 & 0.917 & 0.165 & 15.8 & 84.9 & 0.652 \\ \hline
musk v2 & 0.978 & 0.943 & 0.984 & 0.136 & 4.0 & 87.9 & 0.839 \\ \hline
parkinson & 0.913 & 0.835 & 0.931 & 0.201 & 9.2 & 83.0 & 0.702 \\ \hline
phishing websites & 0.972 & 0.950  & 0.980  & 0.214 & 2.2 & 82.2 & 0.913 \\ \hline
planning relax & 0.698 & 0.578 & 0.695 & 0.293 & 27.5 & 70.5 & 0.460 \\ \hline
rice cammeo  & 0.922 & 0.889 & 0.936 & 0.267 & 3.3  & 77.9 & 0.846 \\ \hline
room occupancy & 0.998 & 0.993 & 0.998 & 0.079 & 0.4 & 93.1 & 0.946 \\ \hline
sonar & 0.818 & 0.695 & 0.853 & 0.274 & 14.3 & 76.0 & 0.605 \\ \hline
speaker accent  & 0.805 & 0.589 & 0.855 & 0.127 & 14.3 & 87.1 & 0.465 \\ \hline
vertebral column & 0.842 & 0.77  & 0.876 & 0.269 & 7.2 & 76.7 & 0.648 \\ \hline
vowel & 0.973 & 0.696 & 0.983 & 0.028 & 9.7 & 97.3 & 0.605 \\ \hline
winequality red & 0.698 & 0.572 & 0.745 & 0.223 & 10.3 & 76.9 & 0.320 \\ \hline \hline
\textbf{Mean}  & \textbf{0.856} & \textbf{0.761} & \textbf{0.877} & \textbf{0.189} & \textbf{8.6} & \textbf{81.9} & \textbf{0.661}\\ \hline
\end{tabular}
\end{center}
\end{table}

The experimental analysis section is designed to evaluate the effectiveness of the probabilistic confusion matrix across various classifiers and datasets. The study involves 21 datasets from the UCI Machine Learning Repository \cite{Dua:2019}, varying in the number of samples (100 to 19,020), number of variables (4 to 166), and number of classes (2 to 15) (see Table \ref{tab:datasets}). The performance of 3--Nearest Neighbors (3--NN), Naïve Bayes (NB), Decision Trees (DT), and Random Forests (RF) classifiers is assessed using both traditional and probabilistic performance measures. Two classification performance measures were selected: accuracy and the MCP curve (Multiclass Classification Performance) \cite{Aguilar2024a,Aguilar2024c}. The choice of the MCP curve is motivated by the fact that this measure is calculated from the probabilities returned by the classifier. Unlike $Acc$ or other popular measures such as F--score or Matthews correlation coefficient, the area under the MCP curve implicitly considers the probabilistic nature of predictions. 

Tables \ref{tab:3NN}, \ref{tab:NB}, \ref{tab:DT} and \ref{tab:RF} show, for each dataset (row), several columns:  Dataset name, accuracy ($Acc$), probabilistic accuracy ($Acc^\star$), certainty accuracy ($Acc_v^\star$), uncertainty accuracy ($Acc_u^\star$), divergence ($div$) in percentage, certainty ratio ($\mathcal{C}_\rho$) in percentage, and the area under the MCP curve (MCP).

The 3--Nearest Neighbor classifier exhibits stable behavior, with divergence of about 4.5, resulting in a small difference between $Acc=0.809$ and $Acc^\star=0.783$. However, the certainty ratio $\mathcal{C}_{3-NN}=73.5\%$ due to the high value of $Acc_u^\star$, i.e., the classifier makes correct predictions by chance based on uncertainty. 

The Naïve--Bayes classifier shows similar behavior to the 3--NN classifier but with generally lower accuracy and uncertainty. The mean accuracy is  $Acc=0.709$, and the certainty ratio is 65.8\%. This suggests that Naïve--Bayes often provides confident predictions, but these predictions are not always as accurate, because the contribution of uncertainty to accuracy is high ($Acc_u^\star$=0.390).

The Random Forest classifier achieves the highest accuracy ($Acc=0.856$), suggesting it is the best classifier. However, it also has a relatively high divergence of 8.6 and a certainty ratio of 81.9\%, indicating that a significant portion of its performance comes from uncertain predictions. This implies that despite high accuracy, RF may not always be the most reliable.

%The Random Forest classifier shows the highest accuracy ($Acc=0.845$), which suggests it is the best classifier. However, the divergence is about 7.7\% and the certainty ratio $\mathcal{C}_{RF}$ is 92.4\%. This 7.6\% uncertainty makes the Random Forest classifier less reliable. 

%In contrast, the Decision Tree classifier presents a 98\% certainty ratio, making it very robust, with IMCP results similar to 3--NN or even better than Naïve--Bayes and Random Forest.

In contrast, the Decision Tree classifier exhibits a very high certainty ratio of 96.2\%, because the contribution of uncertainty to accuracy is very low ($Acc_u^\star=0.058$) indicating robust and reliable performance. Although its accuracy ($Acc=0.799$) is lower than 3--NN and RF, its almost inexistent divergence and very high certainty ratio make it a more dependable classifier. The Decision Tree’s results for the area under the MCP curve are better than those of 3--NN, Naïve--Bayes and Random Forest, underscoring its reliability.

The IMCP curve, which considers probabilistic predictions, consistently shows values closer to $Acc^\star$ than to $Acc$, supporting the argument that probabilistic measures offer a more realistic assessment of classifier performance.

The study, summarized in Table \ref{tab:classifiers}, reveals that classifiers with high accuracy do not necessarily provide reliable predictions when uncertainty is taken into account. For instance, while Random Forest achieves the highest accuracy, its relatively high uncertainty undermines its reliability. The importance of the certainty ratio becomes evident as a critical factor in assessing classifier reliability. Classifiers like Decision Trees with a high certainty ratio demonstrate that accurate and certain predictions are more valuable than high but uncertain accuracy.

In summary, relying solely on accuracy is not advisable when the certainty ratio is not high (close to 100\%). More broadly, we should avoid relying on any classification performance measure when the certainty ratio indicates a significant contribution of uncertainty in the confusion matrix. Therefore, it is more appropriate to analyze the probabilistic confusion matrix to assess the reliability of a performance measure, guided by the certainty ratio $\mathcal{C}_\rho$. This approach ensures that the evaluation of a classifier accounts not only for overall performance but also for the level of confidence and certainty in its predictions. This aspect is critical in many application realms, e.g., in clinical diagnosis.

\begin{table}[t]
\caption{Average results for 3--Nearest Neighbors, Naïve--Bayes, Decision Trees and Random Forests. }
\label{tab:classifiers}
\begin{center}
\fontsize{6}{8}\selectfont
\begin{tabular}{|l|r|r|r|r|r|r|r|}
\hline
Classifier & $Acc$ & $Acc^\star$ & $Acc_v^\star$ & $Acc_u^\star$ & div & $C_\rho$ & IMCP\\ \hline \hline
3--NN & 0.809 & 0.783 & 0.829 & 0.304 & 4.9 & 73.5 & 0.720\\ \hline
Decision Tree & 0.799 & {\color{blue}\textbf{0.799}} & 0.800 & {\color{blue}\textbf{0.058}} & {\color{blue}\textbf{0.1}} & {\color{blue}\textbf{96.2}} & {\color{blue}\textbf{0.757}} \\ \hline
Naive Bayes & {\color{red}\textbf{0.709}} & {\color{red}\textbf{0.689}} & {\color{red}\textbf{0.723}} & {\color{red}\textbf{0.390}} & 4.3 & {\color{red}\textbf{65.8}} & {\color{red}\textbf{0.652}} \\ \hline
Random Forest & {\color{blue}\textbf{0.856}} & 0.761 & {\color{blue}\textbf{0.877}} & 0.189 & {\color{red}\textbf{8.6}} & 81.9 & 0.661\\ \hline
\end{tabular}
\end{center}
\end{table}

\section{Conclusions}\label{sec:conclusions}

The experimental analysis supports the need for probabilistic evaluation measures to accurately reflect the reliability of classifiers. The results suggest that classifiers with lower divergence and higher certainty ratio may often be preferable in applications where reliability is crucial, even if their traditional accuracy is not the highest. Decision Trees stands out as the most reliable classifier due to its high certainty ratio, making it a strong candidate for applications where reliability and interpretability are critical.

Decomposing the probabilistic confusion matrix into certainty ($V$) and uncertainty ($U$) matrices provides deeper insights into classifier behavior. For example, the decomposition helps identify how much of the classifier’s performance is attributed to true certainty versus misleading uncertainty. This decomposition is particularly valuable in critical applications like medical diagnosis, where understanding the certainty behind predictions can directly impact decision--making.

%Beyond medicine, the certainty ratio is highly relevant in safety--critical applications such as autonomous driving, aerospace, and industrial automation. For example, in autonomous vehicle decision--making, a high certainty ratio when identifying obstacles (e.g., distinguishing a pedestrian from a shadow) ensures that the vehicle’s actions are based on reliable information, enhancing safety and performance.

The certainty ratio $C_\rho$ is designed to quantify the contribution of certain and uncertain predictions to any classification performance measure that can be derived from a confusion matrix. One of the key advantages of the certainty ratio is its universal applicability. Regardless of the performance metric in use, $C_\rho$ can be seamlessly integrated to provide deeper insights into the reliability of classifier predictions. This makes $C_\rho$ not only a novel metric but also a versatile addition to the toolkit of performance evaluation methods, applicable across any context where classifier reliability matters.

The certainty ratio $C_\rho$ emerges as a critical factor in evaluating classifier reliability. A high certainty ratio indicates that the classifier’s predictions are largely driven by certain and accurate decisions, enhancing its trustworthiness. This insight is especially valuable in fields such as medicine, finance, and safety--critical systems, where the cost of uncertainty can be high. In clinical settings, the certainty ratio can serve as a key indicator of how much a diagnostic model’s performance is driven by confident predictions versus uncertain ones. For example, in predicting the likelihood of a patient having a particular type of cancer, a high certainty ratio would indicate that the model’s predictions are reliable and likely based on strong diagnostic signals. In contrast, if the certainty ratio is low, clinicians might interpret the diagnostic model’s output with caution, recognizing that the predictions are influenced by ambiguous or conflicting evidence. 

The certainty ratio  $C_\rho$ can also inform the iterative development of machine learning models. For instance, models with consistently low certainty ratios may require additional features, more training data, or refined algorithms to enhance the certainty of their predictions. Future research could focus on developing classifiers that optimize not only for a specific performance metric but also for certainty, explicitly minimizing the contribution of uncertain predictions to overall performance.

Another potential research direction consists in integrating the certainty ratio into explainability frameworks (e.g., SHAPLEY values \cite{Shapley1953} or LIME \cite{Ribeiro20161135}) to provide end--users with a clearer understanding of when and why a model is certain or uncertain about its predictions.

\section*{Acknowledgment}

This work was supported by Grants PID2020-117759GB-I00 and PID2023-152660NB-I00 funded by the Ministry of Science, Innovation and Universities.

\bibliographystyle{unsrt}  
%\bibliography{references}  %%% Remove comment to use the external .bib file (using bibtex).
%%% and comment out the ``thebibliography'' section.
\bibliography{bibliography}

%%% Comment out this section when you \bibliography{references} is enabled.
%\begin{thebibliography}{1}

%\end{thebibliography}

\end{document}